\documentclass[11pt]{article}
\usepackage{consilr2024}
\usepackage{times}
\usepackage{url}
\usepackage{latexsym}
\usepackage{authblk}
\usepackage{graphicx}
\usepackage{booktabs}
\usepackage{tabularx}
\usepackage{hyperref}
\usepackage{listings}
\usepackage{xcolor}
\usepackage{amsmath,amssymb}

\definecolor{codegreen}{rgb}{0,0.6,0}
\definecolor{codegray}{rgb}{0.5,0.5,0.5}
\definecolor{codepurple}{rgb}{0.58,0,0.82}
\definecolor{backcolour}{rgb}{0.95,0.95,0.92}
\lstdefinestyle{json}{
    backgroundcolor=\color{backcolour},
    commentstyle=\color{codegreen},
    keywordstyle=\color{magenta},
    stringstyle=\color{codepurple},
    basicstyle=\ttfamily\footnotesize,
    breaklines=true,
    captionpos=b,
    keepspaces=true,
    showspaces=false,
    showstringspaces=false,
    showtabs=false,
    tabsize=2,
    frame=single,
    rulecolor=\color{black!30}
}

\title{\textbf{Do LLMs Understand Romanian Driving Laws? A Study on Multimodal and Fine-Tuned Question Answering}}

\author[1]{Eduard Barbu}
\author[1]{Adrian Marius Dumitran}
\affil[1]{Faculty of Mathematics and Informatics, University of Bucharest, Romania\\
\texttt{eduard\_barbu\_13@yahoo.com, marius.dumitran@unibuc.ro}}

\date{}

\begin{document}
\maketitle

\begin{abstract}
Ensuring that both new and experienced drivers master current traffic rules is critical to road safety. This paper evaluates Large Language Models (LLMs) on Romanian driving-law QA with explanation generation. We release a 1{,}208-question dataset (387 multimodal) and compare text-only and multimodal SOTA systems, then measure the impact of domain-specific fine-tuning for Llama 3.1-8B-Instruct and RoLlama 3.1-8B-Instruct. SOTA models perform well, but fine-tuned 8B models are competitive. Textual descriptions of images outperform direct visual input. Finally, an LLM-as-a-Judge assesses explanation quality, revealing self-preference bias. The study informs explainable QA for less-resourced languages.

\end{abstract}


\section{Introduction}
Adherence to traffic regulations is critical for the safety of all road users. A key component of driver education is mastering the theoretical knowledge, which is typically assessed through multiple-choice exams. In Romania, these exams feature both text-only questions and questions that require interpreting images of traffic signs, intersections, or complex scenarios. This multimodality presents a significant challenge for modern Artificial Intelligence, requiring models to not only process text but also to correlate visual information with textual context coherently.

Beyond simply choosing the correct answer, an effective learning tool should also be able to explain the reasoning behind the rules. This capability for explainable AI (XAI) is crucial for building trust and enhancing the educational value of AI-powered systems. This paper explores the capabilities of Large Language Models (LLMs) to both answer questions and provide logical explanations in the specialized domain of Romanian road legislation. While LLMs have demonstrated remarkable performance on general benchmarks \cite{b12}, their effectiveness in niche, non-English domains remains less explored.

Our work addresses this gap by focusing on the specialized Question Answering (QA) task of answering driving exam questions. We make the following contributions:
\begin{enumerate}
    \item We introduce a new, comprehensive dataset of 1,208 Romanian driving exam questions, designed to test both correctness and the ability to generate explanations.
    \item We conduct a comparative analysis of several State-of-the-Art (SOTA) LLMs on their QA and explanation generation capabilities, including Gemini 2.5 Flash \cite{b8}, Llama 4 Maverick \cite{b15}, Qwen3 \cite{b4}, and DeepSeekV3 \cite{b9}.
    \item We investigate the impact of domain-specific fine-tuning on the performance of Llama 3.1 \cite{b14} and its Romanian-centric counterpart, RoLlama 3.1 \cite{b17}.
    \item We analyze the performance difference between providing models with images directly versus providing detailed textual descriptions, and we use an LLM-as-a-Judge to evaluate the quality of the generated explanations.
\end{enumerate}

Although the focus of other related work is on general legal question answering, such as using a DeBERTa-based model with bilinear attention and vector similarity for statute alignment \cite{b18}, our study addresses the specific context of Romanian road laws and practical driving test scenarios, exploring the real-world feasibility of using large language models for regulatory compliance, licensing preparation, and context-aware multimodal question answering in a specialized domain.

Unlike general-purpose legal QA sets such as LegalBench and traffic-specific sets like TrafficQA and mLEX-Fr, our dataset directly targets Romanian driving exam questions with multimodal items, bridging text and image context. This complements prior work by adding a non-English, regulatory-domain benchmark with practical licensing relevance. 

The results of this study are relevant for developing intelligent tutoring systems and contribute to a broader understanding of LLM performance for specialized, explainable QA tasks in Romanian.

\section{Methodology}

\subsection{Dataset Creation and Preprocessing}
The primary data source for our work is the official question pool for the Romanian category B driving license exam. Since the official portal of the DRPCIV (Direcția Generală Permise de Conducere și Înmatriculări) \cite{b7} only offers exam simulations without a complete question archive, we utilized a comprehensive public resource, \textit{scoalarutiera.ro} \cite{b22}, which aggregates these official questions.

Using Python with the Selenium library \cite{b20}, we programmatically extracted 1,208 unique questions. 387 questions include images, while the remaining 821 are text-only (32.0\% vs. 68.0\%). This distribution reflects the format of the real world exam, but it results in fewer multimodal samples for training. For each question, we collected the following fields: the question text, three multiple-choice options (A, B, C), the correct answer(s), a detailed explanation for the correct answer, and, if present, a URL to the associated image. 

A key aspect of our work was handling the multimodal nature of the data. We created two versions of the dataset:
\begin{enumerate}
    \item \textbf{Text-Only Dataset:} For models without visual capabilities, or to test text-based reasoning, we manually created detailed descriptions for every image. These descriptions were carefully crafted to be objective and informative, detailing road signs, markings, vehicle positions, and intended maneuvers, without giving away the answer.
    \item \textbf{Multimodal Dataset:} For vision-language models, we used the original images directly.
\end{enumerate}

\begin{figure}[t]
\label{fig:sample_data}
\centering
\begin{minipage}[c]{0.71\columnwidth}
\begin{lstlisting}[style=json]
{
  "question": "What is the right-of-way rule when encountering this sign?",
  "answers": [
    "A: The left-priority rule",
    "B: The first-to-arrive rule",
    "C: The right-priority rule"
  ],
  "image_description": "A triangular warning sign with a red border, white background, and a black 'X' in the middle. It indicates an uncontrolled intersection of roads with equal priority.",
  "correct_answers": ["C"],
  "explanation": "The sign, named 'Intersection of roads,' warns of an upcoming uncontrolled intersection. At such intersections, you must yield to vehicles approaching from your right, applying the right-priority rule."
}
\end{lstlisting}
\end{minipage}
\hfill 
\begin{minipage}[c]{0.27\columnwidth}
\centering
\includegraphics[width=\linewidth]{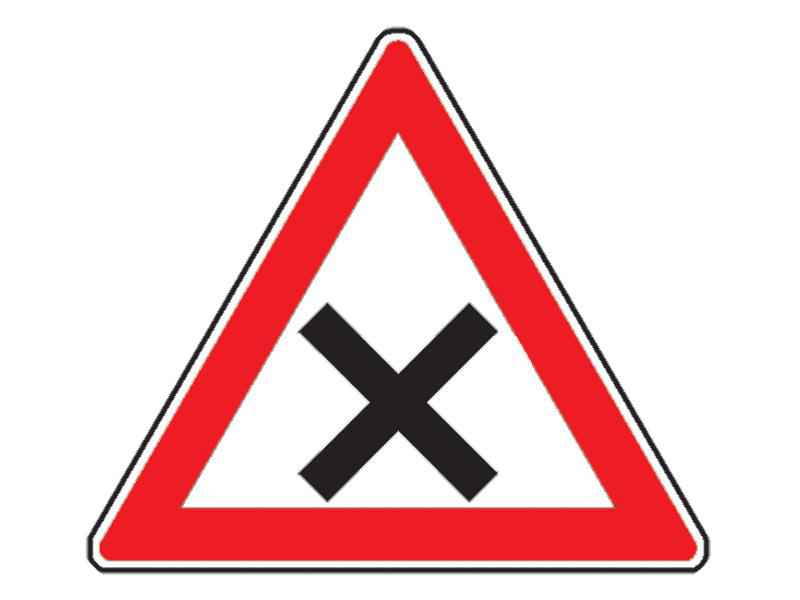}

\end{minipage}

\caption{An example from our dataset, translated for clarity. It tests the model's understanding of the 'Intersection with a road of equal priority' sign (right) and the corresponding 'right-priority rule.' The textual data (left) provides all necessary context for text-only models.}
\label{fig:sample_data_intersection}
\end{figure}

The final datasets were structured in JSONL format for efficient processing \cite{b19}. split\href{https://huggingface.co/datasets/AnonymousDrive/RoDriveImages}{All images} present and \href{https://huggingface.co/datasets/AnonymousDrive/RoDrive}{the base dataset} were made publicly available on Hugging Face. The datasets are distributed under the Hugging Face CC BY-SA 4.0 license and comply with GDPR requirements for user data. We intentionally do not translate the dataset into English, as the questions are specific to Romanian driving laws and their meaning does not generalize meaningfully to other languages or legal contexts. A sample entry from the dataset is shown in Fig.~\ref{fig:sample_data}.

\subsection{Models}
We selected a diverse range of models to cover different architectures, sizes, and capabilities.
\begin{itemize}
    \item \textbf{Gemini 2.5 Flash} \cite{b8}: A fast and powerful multimodal model from Google, used both for evaluation and as our LLM-judge.
    \item \textbf{Llama 4 Series} \cite{b15}: Meta's latest generation of models. We tested Llama 4 Maverick (400B parameters) and Llama 4 Scout (109B parameters), both of which are multimodal.
    \item \textbf{Qwen3-235B-A22B} \cite{b4}: A high-performing multilingual model from Alibaba Cloud.
    \item \textbf{DeepSeekV3} \cite{b9}: A model from DeepSeek AI known for its strong coding and reasoning abilities.
    \item \textbf{Llama 3.1-8B-Instruct} \cite{b14}: A popular open-source model from Meta, serving as a baseline for our fine-tuning experiments.
    \item \textbf{RoLlama 3.1-8B-Instruct} \cite{b17}: A version of Llama 3.1 further pre-trained on Romanian corpora by the OpenLLM-Ro community, designed for better performance on Romanian language tasks.
\end{itemize}
    
\subsection{Evaluation Protocol}
We designed a comprehensive evaluation protocol to assess model performance across multiple dimensions.

\subsubsection{Accuracy on Question Answering}
The primary task was multiple-choice QA. For each question in the test set, models were prompted to first provide the letter(s) of the correct answer(s) followed by a brief explanation. Our main performance metric is \textbf{accuracy}, calculated as the percentage of questions where the model identified all correct answer letters perfectly. This part of the research utilized \~40 USD in order to process all prompts for all models using their respective API.

\subsubsection{Qualitative Evaluation using LLM-as-a-Judge}
To assess the quality of the generated explanations, we employed an LLM-as-a-Judge methodology \cite{b27}. We designated the top-performing model, Gemini 2.5 Flash, as the evaluator. Gemini was tasked to assign a quality score from 0 to 5 to each response based on the correctness, coherence, and relevance of the explanation.

\subsection{Fine-tuning Procedure}
To investigate the impact of domain adaptation, we fine-tuned two open-source models: the general-purpose Llama 3.1-8B-Instruct \cite{b14} and the Romanian-centric RoLlama 3.1-8B-Instruct \cite{b17}. The fine-tuning was performed on our text-only training set using the Hugging Face `transformers` \cite{b25} and `datasets` \cite{b13} libraries. Key hyperparameters included a learning rate of 2e-4, a maximum of 200 training steps, and a `gradient\_accumulation\_steps` of 4 to simulate a larger batch size. All fine-tuning experiments were conducted in Google Colab using an NVIDIA L4 GPU (24GB VRAM), thus using only \~10 USD for this experiment.
\section{Results and Analysis}

Our experiments were designed to answer three primary questions: 1) How do state-of-the-art (SOTA) models perform on this specialized domain out-of-the-box? 2) Can domain-specific fine-tuning elevate the performance of smaller, open-source models? 3) How do modern multimodal models handle direct visual input versus curated textual descriptions?

\subsection{SOTA Model Performance on Text-Only Data}
\begin{figure}[t]
\centerline{\includegraphics[width=\columnwidth]{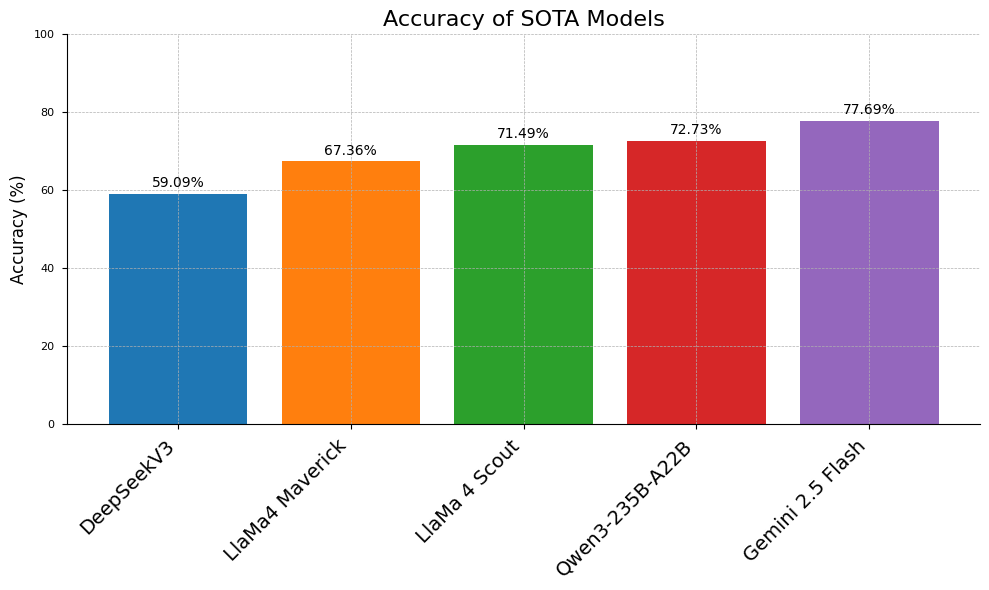}}
\caption{Accuracy of SOTA models on the text-only test set.}
\label{fig:sota_accuracy}
\end{figure}

We first established a performance baseline by evaluating five leading SOTA models on our text-only test set. The results, visualized in Fig.~\ref{fig:sota_accuracy} and detailed in Table~\ref{tab:sota_results}, reveal a clear performance hierarchy.

Gemini 2.5 Flash emerged as the most capable model, achieving an accuracy of \textbf{77.69\%}. It was followed by Qwen3-235B-A22B (72.73\%) and Llama 4 Scout (71.49\%), which both demonstrated strong, albeit lower, performance. Interestingly, the larger Llama 4 Maverick (67.36\%) was outperformed by its smaller counterpart, Llama 4 Scout, suggesting that Scout's architecture may be better optimized for this type of focused QA task. DeepSeekV3 lagged significantly with an accuracy of 59.09\%, indicating that its general knowledge base is not as well-aligned with the specifics of Romanian legislation.

To assess whether the observed accuracy gaps are statistically significant, we ran McNemar’s test on pairwise model predictions. Representative results are listed in Table~\ref{tab:mcnemar_results}. Whereas \textbf{DeepSeek-v3} differs significantly from every other model (\(p < 0.001\)), the comparison between \textbf{Llama 4 Scout} and \textbf{Qwen 3--235B--A22B} shows no significant difference (\(p > 0.05\)), suggesting that the two models have similar decision boundaries and largely overlapping error patterns.

\begin{table}[htbp]
\caption{SOTA Model Performance on the Text-Only Dataset}
\begin{center}
\begin{tabular}{@{}lcc@{}}
\toprule
\textbf{Model} & \textbf{Accuracy (\%)} & \textbf{Judge Score (Avg)} \\
\midrule
Gemini 2.5 Flash & \textbf{77.69} & \textbf{4.53} \\
Qwen3-235B-A22B & 72.73 & 4.24 \\
Llama 4 Scout & 71.49 & 4.05 \\
Llama 4 Maverick & 67.36 & 3.96 \\
DeepSeekV3 & 59.09 & 3.90 \\
\bottomrule
\end{tabular}
\label{tab:sota_results}
\end{center}
\end{table}

\begin{table}[htbp]
\centering
\caption{Examples of McNemar's Test Results for Model Comparisons}
\label{tab:mcnemar_results}
\begin{tabular}{|l|l|l|l|}
\hline
\textbf{Model A} & \textbf{Model B} & \textbf{Statistic} & \textbf{Significant} \\ \hline
DeepSeekV3 & Gemini 2.5 Flash & 24.5063 & Yes ($p$ = 0.0000) \\ \hline
DeepSeekV3 & Llama4 Maverick  & 4.8784  & Yes ($p$ = 0.0272) \\ \hline
Gemini 2.5 Flash & Llama4 Maverick & 8.3478 & Yes ($p$ = 0.0039) \\ \hline
Llama 4 Scout   & Qwen 3235ba22b   & 0.0656 & No ($p$ = 0.7979) \\ \hline
\end{tabular}
\end{table}

\subsection{Impact of Fine-Tuning on Open-Source Models}
A key hypothesis of our work was that domain-specific fine-tuning could make smaller, more accessible models competitive. We tested this by fine-tuning Llama 3.1-8B and its Romanian-centric counterpart, RoLlama 3.1-8B.

\begin{figure}[htbp]
\centerline{\includegraphics[width=\columnwidth]{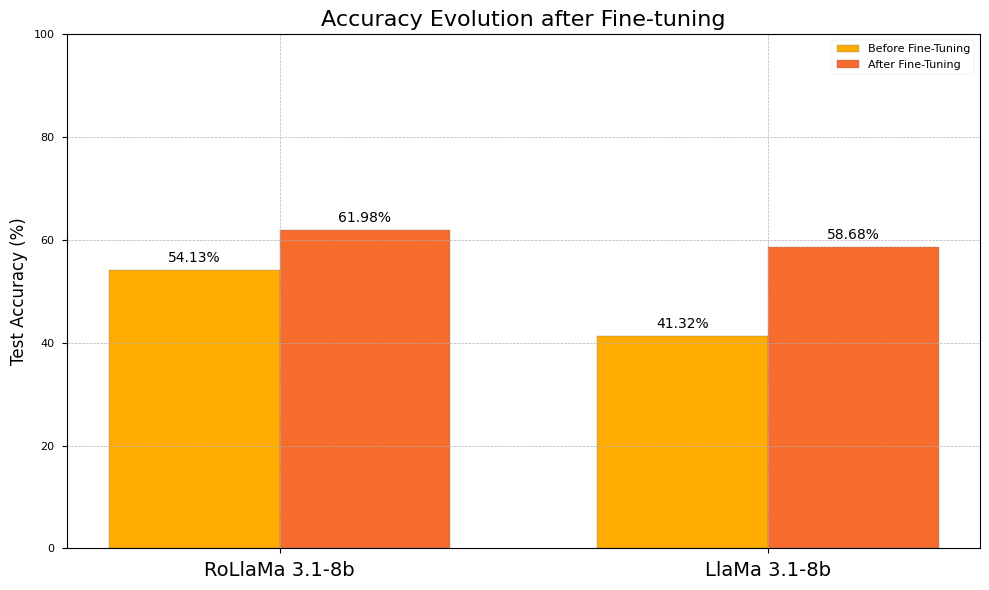}}
\caption{Accuracy improvement from fine-tuning.}
\label{fig:finetuning_chart}
\end{figure}

The results, presented in Table~\ref{tab:finetuning_results} and visualized in Fig.~\ref{fig:finetuning_chart}, confirm this hypothesis emphatically. The base Llama 3.1 model saw its accuracy leap from 41.32\% to 58.68\%---an improvement of over 17 percentage points. RoLlama 3.1, which already benefited from its Romanian pre-training (starting at 54.13\%), improved further to a final accuracy of \textbf{61.98\%}. This is a critical finding: after fine-tuning on a modest dataset, the 8-billion-parameter RoLlama 3.1 model surpassed the out-of-the-box performance of the much larger DeepSeekV3. This highlights that for specialized domains, targeted data adaptation is a highly effective strategy.

\begin{table}[htbp]
\caption{Performance Before and After Fine-Tuning}
\begin{center}
\begin{tabular}{@{}lcccc@{}}
\toprule
\textbf{Model} & \multicolumn{2}{c}{\textbf{Accuracy (\%)}} & \multicolumn{2}{c}{\textbf{Judge Score}} \\
\cmidrule(lr){2-3} \cmidrule(lr){4-5}
& Before & After & Before & After \\
\midrule
Llama 3.1-8B & 41.32 & 58.68 & 3.05 & 3.77 \\
RoLlama 3.1-8B & 54.13 & \textbf{61.98} & 3.47 & \textbf{3.88} \\
\bottomrule
\end{tabular}
\label{tab:finetuning_results}
\end{center}
\end{table}

\subsection{Image-based vs. Description-based QA}
To understand the current state of multimodal reasoning, we compared model performance when processing images directly versus reading our manually created text descriptions. Table~\ref{tab:multimodal_results} shows that all models performed significantly better when provided with textual descriptions. The performance drop when using direct images was substantial, ranging from 7\% for Gemini to a stark 15\% for Llama 4 Scout. We attribute this gap to the low resolution and compression artifacts of the source images and the inherent difficulty for models to interpret nuanced visual details correctly. This suggests that for this task, a high-quality textual description is currently more reliable than direct visual input. 

\begin{table}[htbp]
\caption{Accuracy: Direct Image vs. Text Description}
\begin{center}
\begin{tabular}{@{}lcc@{}}
\toprule
\textbf{Model} & \textbf{Accuracy (Text Desc.)} & \textbf{Accuracy (Image)} \\
\midrule
Gemini 2.5 Flash & \textbf{77.69\%} & \textbf{70.66\%} \\
Llama 4 Scout & 71.49\% & 56.61\% \\
Llama 4 Maverick & 67.36\% & 59.51\% \\
\bottomrule
\end{tabular}
\label{tab:multimodal_results}
\end{center}
\end{table}

\subsection{Analysis of Explanation Quality}
\begin{figure}[t]
\centerline{\includegraphics[width=\columnwidth]{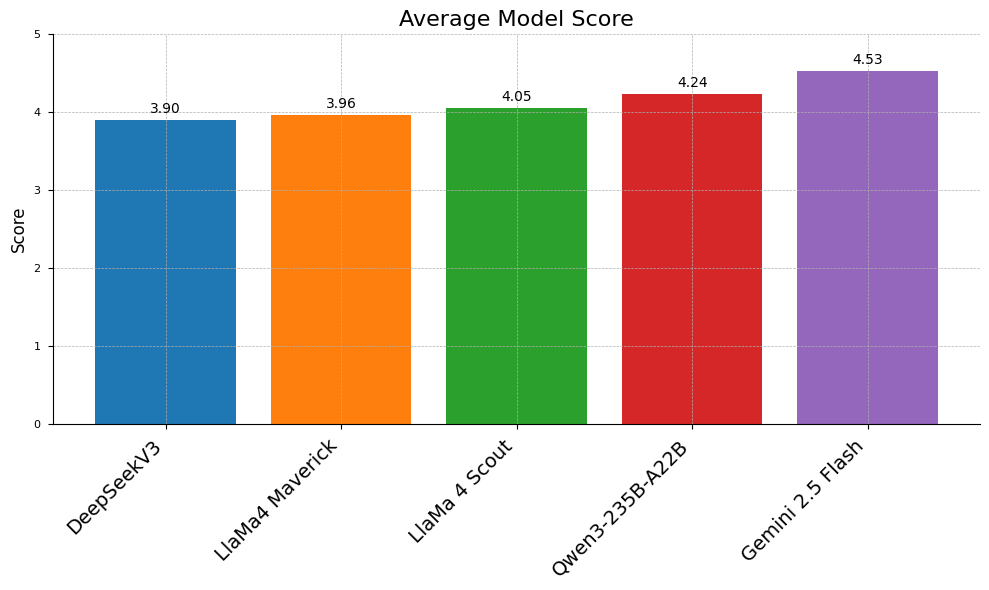}}
\caption{Average explanation quality score (0-5) for SOTA models, evaluated by the Gemini 2.5 Flash LLM-as-a-Judge.}
\label{fig:judge_scores}
\end{figure}

Beyond correctness, we evaluated the quality of the generated explanations using Gemini 2.5 Flash as an LLM-as-a-Judge. The average explanation scores, presented in Fig.~\ref{fig:judge_scores}, show a strong positive correlation with accuracy. Models that were more accurate also tended to provide better-rated explanations.

An important observation from this evaluation is the evidence of self-preference bias. Gemini 2.5 Flash awarded its own explanations the highest score of \textbf{4.53}, significantly higher than the next best model. This highlights a key challenge in LLM-as-a-Judge methodologies and underscores the need for caution when interpreting automated quality scores.

\section{Discussion}
Our experiments provide several key insights into the application of LLMs for specialized domains.

First, there is a clear performance gap between general SOTA models and those specifically adapted for a domain. The dramatic improvement from fine-tuning demonstrates that specialization is a highly effective strategy, especially for less-resourced languages like Romanian. A relatively small, well-curated dataset can elevate a modest open-source model to be competitive with much larger, general-purpose models, showcasing a cost-effective path to high performance.

Second, the current state of multimodality for this specific task shows that "a good description is worth a thousand pixels." The performance degradation when using images directly points to ongoing challenges in visual perception, particularly with low-quality or nuanced visual data. This has important implications for the design of practical applications, suggesting that a hybrid approach, using vision models to generate initial descriptions that are then refined or used by a text model, may be optimal until visual interpretation capabilities mature further.

Finally, our use of an LLM-as-a-Judge, while efficient, revealed a tangible self-preference bias. Gemini consistently scored its own answers higher than those of other models, even when accuracies were comparable. This serves as a cautionary tale: automated evaluation metrics, even sophisticated ones, should be used with care and ideally validated with human expert judgment. The scores attributed by the judge were, however, strongly correlated with accuracy, indicating their utility as a directional measure of quality.




\section{Conclusion and Future Work}
In this paper, we evaluated the ability of modern LLMs to understand Romanian driving legislation through a comprehensive set of experiments on a newly created dataset. We demonstrated that SOTA models like Gemini 2.5 Flash can achieve high accuracy, but that fine-tuning smaller, language-specific models like RoLlama 3.1 offers a path to competitive performance in niche domains. Our analysis of multimodal capabilities revealed that detailed textual descriptions currently yield better results than direct image processing for this task, highlighting the importance of data quality and the challenges of visual reasoning.

The primary limitations of this work are the quality of the source images and the potential subjectivity in the manual image descriptions. Future work could focus on using generative models to enhance or standardize the images.

The methodology and findings presented here can be extended to other specialized domains requiring multimodal reasoning, such as medical chart interpretation or technical diagram analysis. The fine-tuned models and datasets we developed can serve as a foundation for building practical applications, such as interactive, personalized learning systems for aspiring drivers in Romania, ultimately contributing to better education and safer roads. The code and training scripts are available on \href{https://anonymous.4open.science/r/RoDrive-62CF}{GitHub}. Our results are advisory only; final answers must be validated by certified driving instructors.

\bibliographystyle{consilr}
\bibliography{refs}

\end{document}